**SyriSign: A Parallel Corpus for Arabic Text to Syrian Arabic Sign Language Translation**

Mohammad Amer Khalil +, *Raghad Nahas +, Ahmad Nassar +, Khloud Al Jallad*

*Arab International University, Daraa, Syria*

## Abstract

Sign language is the primary approach of communication for the Deaf and Hard-of-Hearing (DHH) community. While there are numerous benchmarks for high-resource sign languages, low-resource languages like Arabic remain underrepresented. Currently, there is no publicly available dataset for Syrian Arabic Sign Language (SyArSL). To overcome this gap, we introduce SyriSign, a dataset comprising 1500 video samples across 150 unique lexical signs, designed for text-to-SyArSL translation tasks. This work aims to reduce communication barriers in Syria, as most news are delivered in spoken or written Arabic, which is often inaccessible to the deaf community. We evaluated SyriSign using three deep learning architectures: MotionCLIP for semantic motion generation, T2M-GPT for text-conditioned motion synthesis, and SignCLIP for bilingual embedding alignment. Experimental results indicate that while generative approaches show strong potential for sign representation, the limited dataset size constrains generalization performance. We will release SyriSign publicly, hoping it serves as an initial benchmark.

**Keywords:** Arabic Sign Language (ArSL), Deaf accessibility, SignCLIP, MotionCLIP, T2M-GPT

## 1. Introduction

The Deaf and Hard-of-Hearing (DHH) community in Syria faces substantial challenges in accessing local news. A study in 2021 (Humanitarian Needs Assessment Programme (HNAP), 2021) estimates that approximately 8% of Syrians over the age of 17 experience partial or complete hearing loss. As the majority of local news is delivered in spoken Arabic, DHH individuals are often excluded unless captions are provided. Moreover, this challenge is compounded by the fact that many members of the community do not achieve full proficiency in written Arabic.

Syrian Arabic Sign Language (SyArSL) is a distinct visual-spatial language with its own unique grammar and syntax, representing a regional dialect within the broader Arabic Sign Language (ArSL) family. Despite its importance, professional SyArSL interpreters remain critically scarce, with only 10 officially registered sworn interpreters in Syria (Akkazeh, 2025). This shortage limits real-time interpretation for live broadcasts and public announcements. Furthermore, digital platforms rarely provide content in SyArSL, forcing deaf Syrians to rely on limited and inconsistently provided news or completely lose their rightful access to local news.

This work addresses these barriers by introducing SyriSign, a novel SyArSL dataset to facilitate automated translation research. Moreover, we investigate three distinct approaches to Arabic text-to-SyArSL translation, evaluating the feasibility, strengths, and limitations of various state-of-the-art architectures. These solutions range from retrieval-based models that learn precise text-to-sign correspondences to generative models capable of synthesizing 3D skeletal animations, as well as hybrid approaches that combine both methodologies.

The main contributions of this paper are as follows:

1. **Dataset Creation:** we present SyriSign[1,2], a novel SyArSL dataset comprising 1,500 video samples across 150 unique lexical signs, addressing the scarcity of resources for Syrian Arabic Sign Language.

---

+ Equal Contribution

[1] https://huggingface.co/datasets/Mohammad-Amer-Khalil/SyriSign

[2] https://github.com/Moham-Amer/SyriSign

2. **Benchmark Evaluation:** We provide a comparative evaluation of three state-of-the-art text-to-SL translation architectures: MotionCLIP, T2M-GPT, and SignCLIP, specifically under fine-tuned settings for low-resource contexts.
3. **Methodological Analysis:** We analyze the trade-offs between retrieval-based and generative methods, establishing a performance baseline for future SyArSL production systems.

## 2. Literature Review

### 2.1. Datasets

There are several distinct types of sign language datasets, categorized by their data format and acquisition method:

- Video-based datasets: Capture full spatiotemporal information, including hand shape, motion trajectories, body posture, and facial expressions e.g., (Koller, Zargaran, Ney, & Bowden, 2018).
- Keypoint/skeleton datasets: Represent signs using extracted coordinates of body, hand, and facial joints, providing a computationally efficient alternative to raw video e.g. (Zelinka & Kanis, 2020).
- Image-based datasets: Consist of individual frames or still images extracted from signing videos, such as ArASL (Latif, Mohammad, Alghazo, AlKhalaf, & AlKhalaf, 2019), which contains 54,049 static images.
- Gloss-annotated datasets: Associate sign segments with textual glosses as intermediate representation e.g. (Camgöz, Hadfield, & Koller, 2018) Serving as a linguistic bridge between sign language and written text.
- Motion capture/3D datasets: Record precise 3D hand and body movements using specialized hardware such as sensor gloves or depth-tracking systems e.g., (Jedlička, Krňoul, Železný, & Müller, 2022).

The efforts made by (Luqman, Mohandes, Mahmoud, & I. Sidig, 2021) , (Vaezi Joze & Koller, 2019) ,and (Li, Rodriguez Opazo, Yu, & Li, 2020) are examined to understand their guidelines and rules in SL dataset creation. Picking a number of different factors to compare the datasets, it is worth noting all these datasets are word-level. As shown in Table 1:

*Table 1 State-of-the-art Datasets*

| Factor | **KArSL** (Luqman, Mohandes, Mahmoud, & I. Sidig, 2021) | **MS-ASL (ASL1000)** (Vaezi Joze & Koller, 2019) | **WLASL (WLASL2000)** (Li, Rodriguez Opazo, Yu, & Li, 2020) |
|---|---|---|---|
| **Unique words** | 502 | 1,000 | 2,000 |
| **Signers** | 3 | 222 | 119 |
| **Word Repetitions** | 150 (50 per signer) | Varies, minimum 11 | Varies, average 10.5 |
| **Total number of samples** | 75,300 | 25,513 | 21,083 |
| **FPS** | 30 FPS | 25.5 FPS (Average) | 25 FPS |
| **Word collection method** | Dictionary-Based | Web-mined + OCR | Curated online ASL resources |
| **Video Format, Modalities, and Annotations** | RGB, Depth, Skeleton joint points | RGB only | RGB only |
| **Video Duration** | - | Average video is 64 frames; total duration is 24 hours and 39 minutes | Average video duration 2.4s, maximum duration 8s |

### 2.2. State-of-the-art Models

In this section, we review state-of-the-art models and their underlying methodologies. Sign Language Production (SLP) approaches generally fall into three categories: Rule-Based, which relies on predefined linguistic rules and mapping dictionaries to translate text into sign sequences; Gloss-Based, which uses an intermediate representation (glosses) to translate spoken language into visual signs, typically via a two-stage pipeline (Text-to-Gloss and Gloss-to-Sign); and Gloss-Free, which utilizes end-to-end architectures to bypass intermediate glosses, directly mapping input text to sign representations.

#### 2.2.1. Rule-Based SLP

Most legacy systems dealt with sign language sentences as images; systems such as that of (Alethary, Aliwy, & Ali, 2022) generate signs from the unified Arab dictionary. Avatar technologies were affordable to work with owing to their low cost in time and space and easy appearance customization.

#### 2.2.2. Gloss-Based SLP

Sign language has its own grammatical structure, syntax and vocabulary, which poses a fine challenge in the fields of Sign Language Recognition (SLR), Sign Language Translation (SLT) and Sign Language Production (SLP). Early studies believed that the best and the most effective way to achieve accurate sign production was by preserving sign language's linguistic identity through glosses. Glosses as intermediate representations serve as a bridge, allowing models to learn specific linguistic characteristics that are otherwise difficult to capture when mapping directly from spoken language. While this approach has been foundational, a recent work using progressive transformers (Saunders, Camgoz, & Bowden, 2020) suggests that glosses may actually introduce information bottlenecks rather than improving. This critique motivated subsequent gloss-free research for the shift toward modern gloss-free architectures.

#### 2.2.3. Gloss-Free SLP

While Gloss-Based SLP provides an efficient framework for translating spoken language into sign poses through symbolic representation, limitations arise as the focus shifts toward producing fluid, human-like signs. Gloss-Based methods often struggle with expanding datasets to accommodate a dynamic sign language vocabulary and tend to suffer from information loss during the glossing stage. In contrast, the Gloss-Free approach represents a significant evolution in both SLT and SLP. By removing the need for intermediate gloss annotations, a broader space for Text-to-Pose (T2P) architectures appears. This shift allows researchers to leverage decades of neural network advancements to design more dynamic systems. For example, (Zifan Jiang, 2024) proposed SignCLIP, a method based on Contrastive Language-Image Pretraining (CLIP) tailored for learning the mapping between sign language poses and text, extending CLIP's contrastive learning to 500 thousand text-video pairs across 41 sign languages from SpreadTheSign, and using MediaPipe to extract pose landmarks instead of raw video, they demonstrated a more scalable, end-to-end approach..

*Table 2 Related Works*

| Year | Model | Gloss-Based | Approach | Key idea | Key Result |
|---|---|---|---|---|---|
| 2024 | SignCLIP (Zifan Jiang, 2024) | No | Retrieval | Text-pose embedding alignment | Recall@1: 0.38 |
| 2023 | T2M-GPT (Zhang, et al., 2023) | No | Generative | Tokenized Motion + GPT | R-Precision Top-1: 0.492; FID: 0.141; |
| 2022 | Rule-Based NLP Translator (Alethary, Aliwy, & Ali, 2022) | Yes | Rule-Based | Morphological + syntactic analysis | Accuracy: ~86%; |

| 2022 | MotionCLIP (Tevet, Gordon, Hertz, Bermano, & Cohen-Or, 2022) | No | Hybrid | CLIP-aligned motion in latent space | R-Precision Top-1: 0.533; FID: 0.047; |
|---|---|---|---|---|---|

## 3. Proposed Methodology

### 3.1. Dataset Creation Methodology

SyriSign dataset was specifically created for translating Arabic text within the domain of local news and public announcements, with a focus on the Syrian dialect of sign language. To ensure linguistic relevance to the local news context, vocabularies were selected based on the frequency of occurrence in a regional newspaper over a six-month period. Figure 1 shows the distribution of word categories, and Figure 2 the distribution of noun topics.

Following the recording methodologies conducted in WLASL (Li, Rodriguez Opazo, Yu, & Li, 2020) and KArSL (Luqman, Mohandes, Mahmoud, & I. Sidig, 2021) datasets, we recorded 150 unique word signs. Each sign was recorded by two signers, with each signer performing it five times, resulting in 1,500 video samples, amounting to approximately three hours of footage. All videos were recorded in RGB format and 60 FPS, with a maximum duration of seven seconds per clip.

Fig. 1. Analysis of Word Categories within the SyriSign Dataset

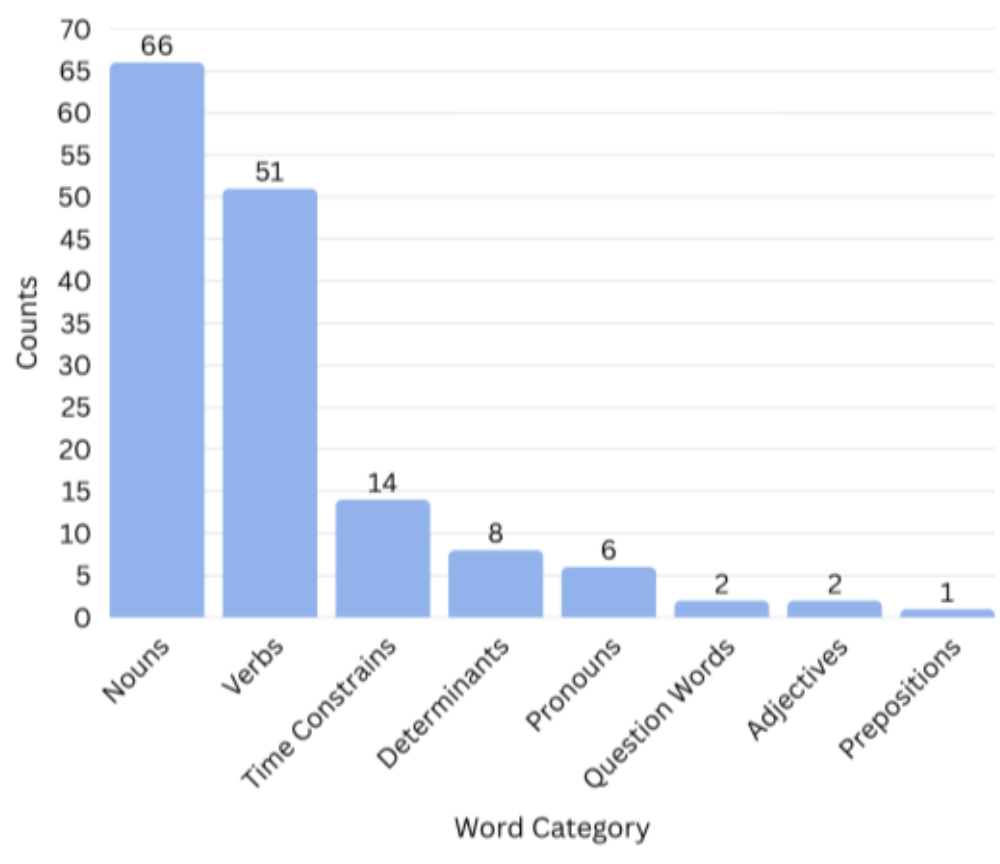

Fig. 2. Noun distribution by topics.

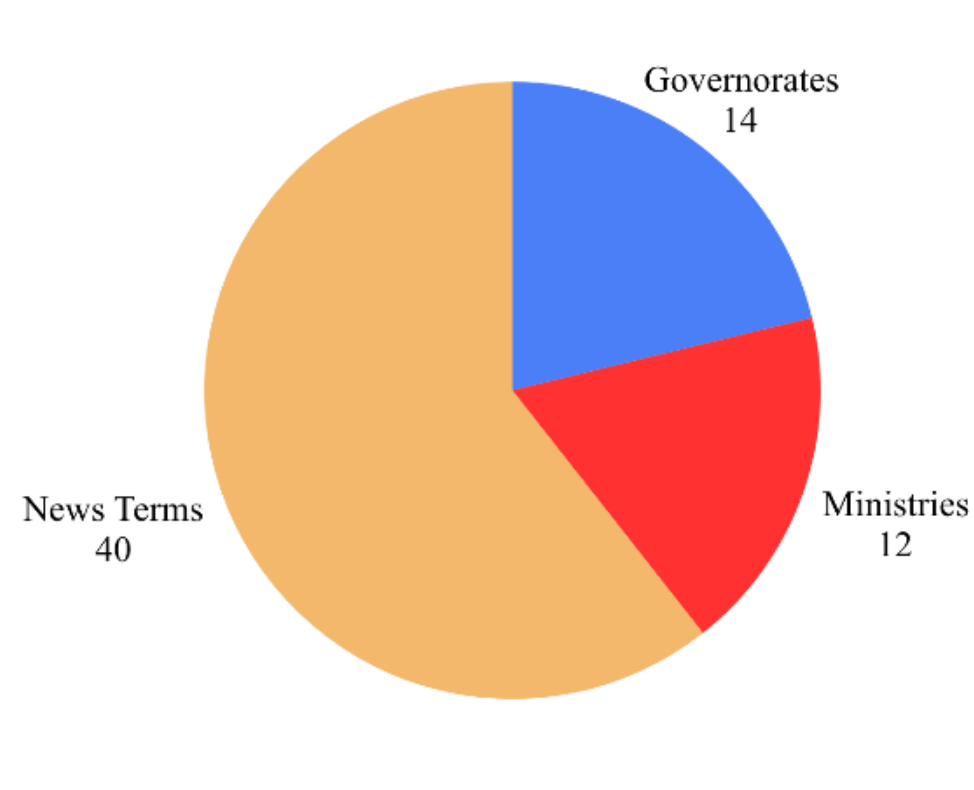

Seven signers, three males and four females, with ranging ages and hearing abilities performed the recorded signs in SyriSign. Figure 3 shows the gender and age distribution of the signers, while Figure 4 shows the distribution of their hearing abilities. The hearing individuals involved include a sworn translator.

Fig. 3. SyriSign number of Signers by age range and gender

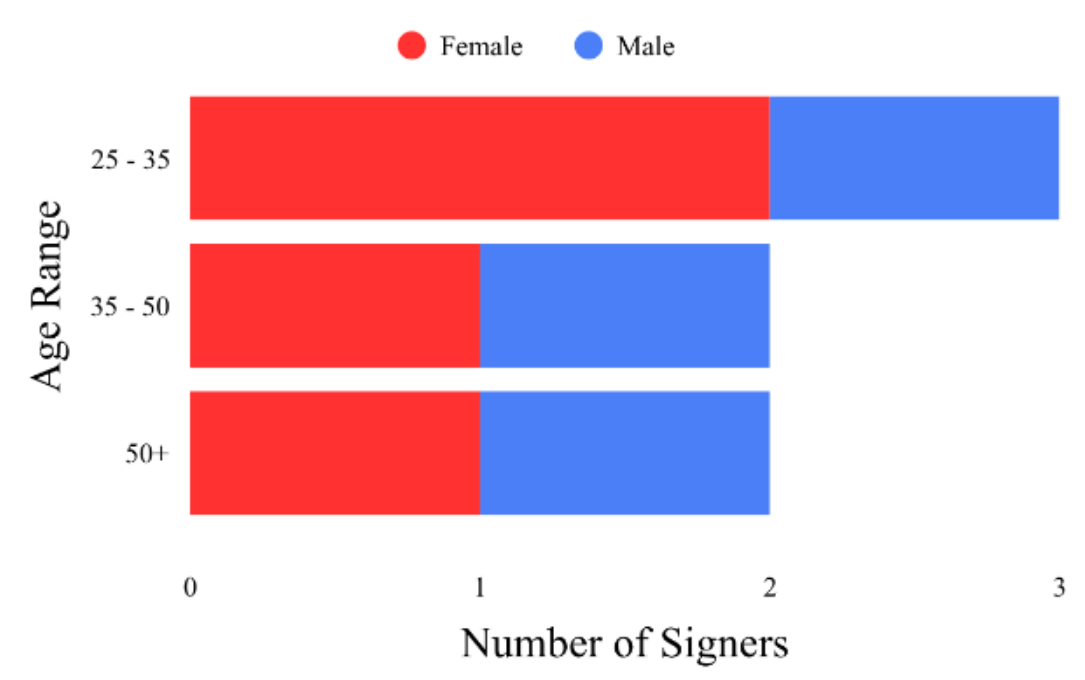

Fig. 4. SyriSign number of contributing signers by hearing abilities

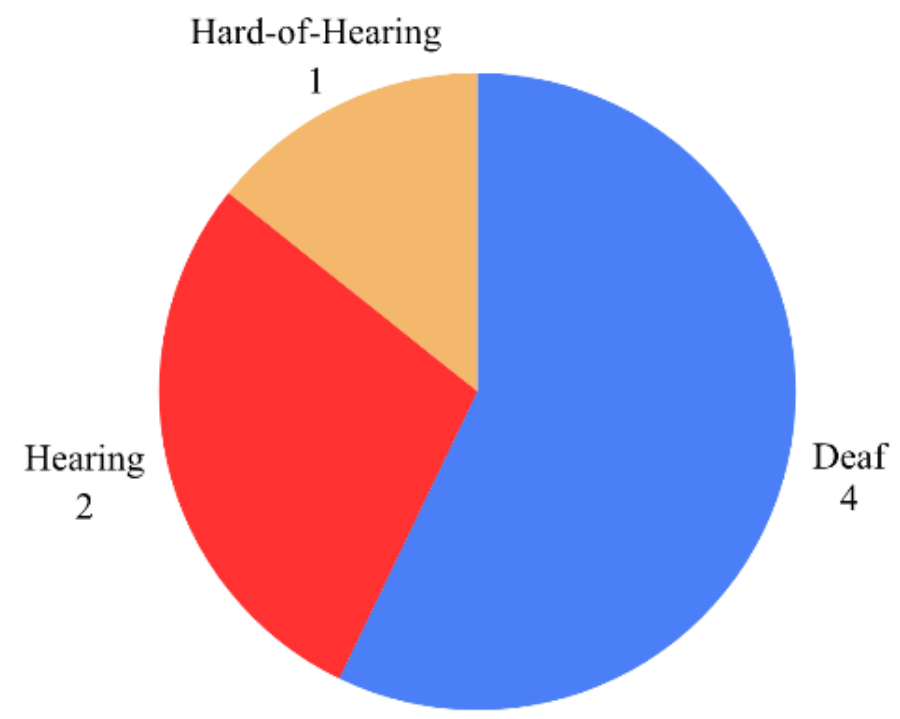

### 3.2. Data Preprocessing

The preprocessing utilized MediaPipe Holistic pipeline (Lugaresi, et al., 2019; Grishchenko & Bazarevsky, 2020) a robust framework developed by Google for integrated full-body landmark detection.

MediaPipe pipeline synchronizes separate models for pose, face, and hand components to track a total of 543 landmarks in real-time. By leveraging this holistic approach, the system captures the facial expressions and hand gestures essential for accurate sign language representation.

### *3.3. Models Architectures*

#### *3.3.1. MotionCLIP*

MotionCLIP (Tevet, Gordon, Hertz, Bermano, & Cohen-Or, 2022) is a 3D human motion auto-encoder designed to align motion latent spaces with the semantic structure of the CLIP model. It utilizes a transformer-based encoder-decoder architecture to reconstruct 3D human Motion Sequences, usually in skeletal form (joint rotations over time) as inputs, while matching corresponding text-image embeddings. The primary objective is to inherit the rich semantic relationships from CLIP's large-scale datasets to enable motion generation from text descriptions.

##### *3.3.1.1. Original Loss Function*

The model optimization is driven by a weighted sum of three primary losses, as shown in Equation 1:

- Motion Reconstruction Loss: Minimizes the deviation between predicted and actual joint rotations.
- Text Alignment Loss: Minimizes the distance between motion and text vectors in the latent space.
- Image Alignment Loss: Aligns motion embeddings with rendered visual representations.

$$L\text{total} = L\text{recon} + \lambda text * L\text{text} + \lambda image * L\text{image} \quad (1)$$

##### *3.3.1.2. Modified MotionCLIP for Arabic Sign Language*

As shown in Figure 5, to adapt MotionCLIP for Arabic Sign Language, our pipeline includes a translation layer to bridge Arabic text with the CLIP text encoder. We replaced the standard input with truncated motion sequences extracted via MediaPipe for better skeletal compatibility. Our implementation prioritizes embedding alignment by focusing on a specific decomposition of the loss function, as shown in Equation 2:

- Reconstruction Loss: Measures difference between predicted and ground-truth joints over all time steps.
- Velocity Smoothness Loss: Ensures smoothness between frames by penalizing inconsistencies in motion velocity.
- Latent Regularization Loss: Constrains the magnitude of the latent motion representation to promote stability and prevent embedding drift.

$$L\text{total} = \lambda recon * L\text{recon} + \lambda vel * L\text{vel} + \lambda latent * L\text{latent} \quad (2)$$

Fig. 5. Modified MotionCLIP Pipeline Diagram

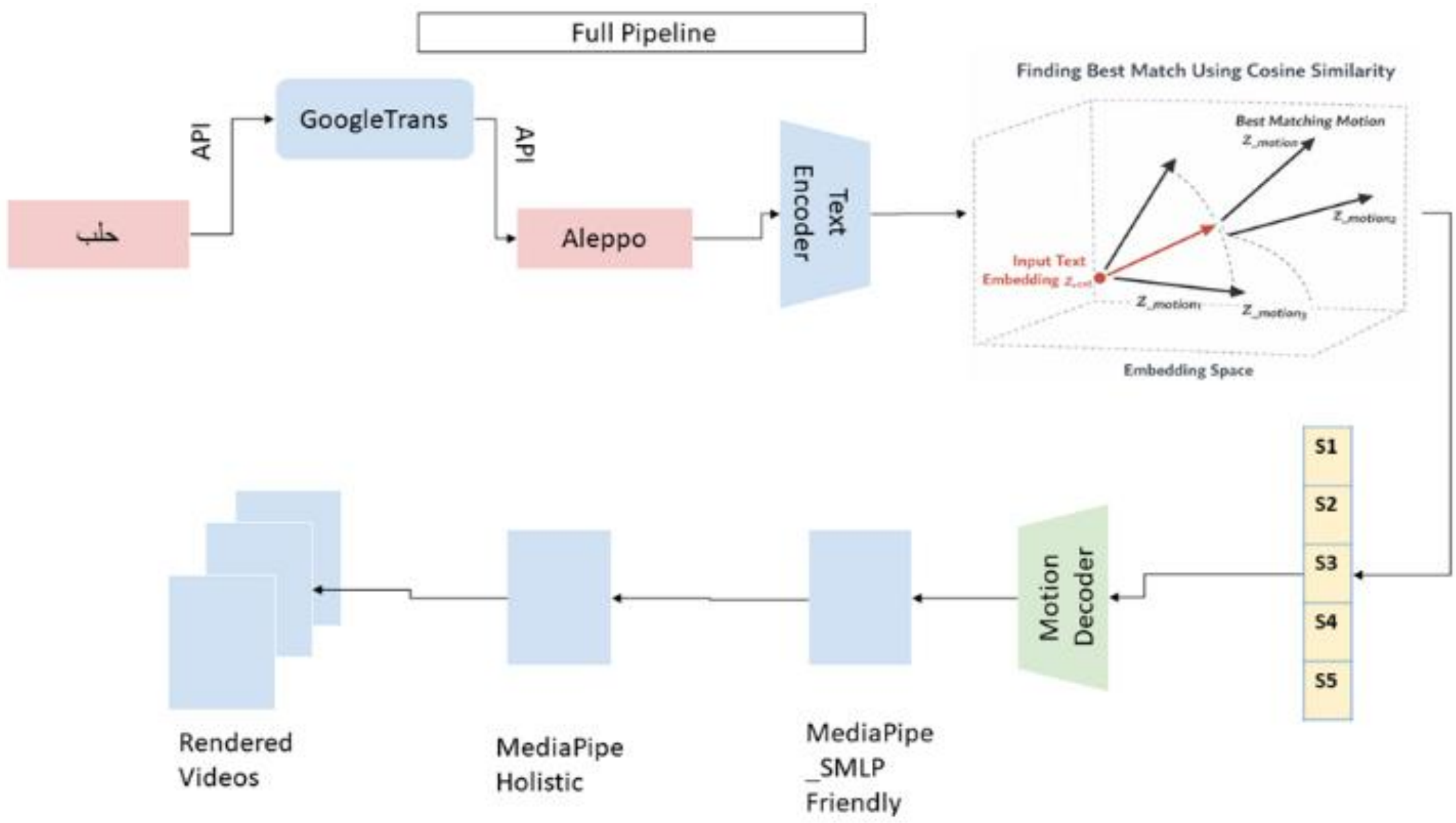

*3.3.2. Text-to-Motion Generative Pretrained Transformer (T2M-GPT):*

T2M-GPT (Zhang, et al., 2023) is a two-stage, text-conditioned framework that synthesizes human motion by decomposing the process into learning discrete representations of motion. It utilizes a Vector Quantized Variational Autoencoder (VQ-VAE) to map continuous motion into a discrete codebook of tokens. Instead of generating continuous motion directly, T2M-GPT predicts sequences of discrete motion tokens from a fixed-size codebook conditioned on text embeddings. These tokens are later decoded using the Generative Pretrained Transformer (GPT) into continuous skeletal motion producing sign language gesture.

*3.3.2.1. Original Loss Function*

The model optimization relies on two distinct stages: VQ-VAE Stage, and GPT Stage.

As for the VQ-VAE loss function, it consists of three main Components shown in Equation 3, as follows:

- Reconstruction Loss minimizes the difference between predicted and ground truth joint points
- Embedding Loss for codebook alignment as it updates the codebook vectors by minimizing the distance between the encoder outputs and their nearest codebook entries
- Commitment Loss for encoder stability, as it forces the encoder output to remain close to the selected codebook vectors.

$$L\text{vq} = L\text{recon} + L\text{embed} + L\text{commit} \quad (3)$$

As for GPT loss function, the model is trained to minimize the Negative Log-Likelihood of the token sequence. Random token corruption is applied during training to improve inference robustness, shown in Equation 4.

$$L\text{trans} = E_{S\sim p(s)}[-\log P(S|c)] \quad (4)$$

*3.3.2.2. Modified T2M-GPT For Arabic Sign Language*

We adapted the framework to create T2M-GPT-ARABIC, an end-to-end pipeline tailored for Arabic Sign Language (ArSL). Key modifications include:

- Data Processing: Full-body key points were extracted using MediaPipe Holistic. Facial landmarks were excluded to reduce dimensionality and improve reconstruction stability in a low-resource setting. However, this simplification removes important non-manual linguistic features, which are essential in sign language. Future work will include full facial features to improve linguistic completeness
- Temporal Windowing: Input sequences were processed using fixed-length windows. We experimented with short (84 frames) and long (168 frames) windows to better capture extended temporal features.
- Text Encoding: We replaced the standard text encoder with AraT5 (Nagoudi, Elmadany, & Abdul-Mageed, 2022), a transformer model pretrained on large-scale Arabic corpora. This ensures the motion generator receives rich, contextually accurate semantic embeddings for the Syrian dialect.

To ensure effective utilization of discrete motion tokens, the VQ-VAE was trained independently using Exponential Moving Average (EMA) and codebook reset mechanisms to prevent codebook collapse. We implemented an Enhanced Reconstruction Loss (Ltotal) as shown in Equation 5:

$$L\text{total} = L\text{pose} + \lambda * L\text{delta} \quad (5)$$

Where Lpose represents the spatial distance between predicted and ground-truth joint positions, Ldelta represents velocity that loss ensures temporal smoothness between consecutive frames, and λ corresponds to the velocity weighting factor, controlling the relative contribution of temporal smoothness. The transformer is optimized using categorical cross-entropy loss over motion tokens. The following figure (Figure 6) shows the pipeline of the Arabic implementation of T2M-GPT.

Fig. 6. T2M-GPT-ARABIC Pipeline

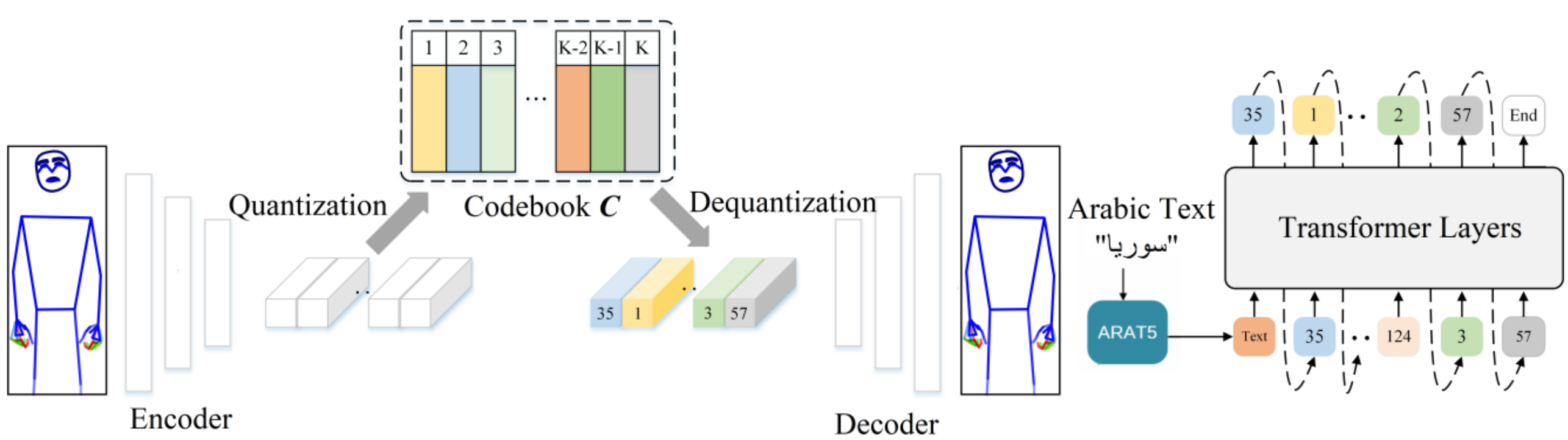

*3.3.3. SignCLIP*

SignCLIP (Jiang, et al., 2024) is a CLIP-style multimodal embedding model. It uses contrastive learning to align written text with sign language video representations in a shared embedding space. It adapts VideoCLIP architecture by replacing general-domain video encoders with sign-language-specific pose estimations derived from MediaPipe Holistic. We fine-tuned it on our created dataset. The model is optimized using a contrastive InfoNCE loss to align text and pose embeddings.

*3.3.3.1. Preprocessing*

To improve computational efficiency and model focus, we implemented a reduction in total landmarks, keeping only the most linguistically relevant points. The original 543 MediaPipe landmarks were filtered down to 193, as follows:

- Pose: Reduced from 33 to 23 landmarks by removing lower-body points (hips, legs, knees, ankles, feet, and heels).
- Face: Simplified from 468 to 128 only, focusing on facial contours.
- Hands: All 42 landmarks (21 per hand) were kept unchanged to capture critical features.

For a video of n frames, the resulting visual encoding is a NumPy array of shape (n,193,3), representing the (x, y, z) coordinates of the optimized landmark set.

*3.3.3.2. Modified AraSignCLIP Pipeline*

The modified pipeline includes Arabic to English translation layer using Open Parallel Corpus Machine Translation (OPUS MT) (Aulamo, et al., 2023) . The pipeline generates a SignCLIP-style prompt consisting of ISO 639-3 codes of the text (source) and sign (target) languages, followed by the text itself. The ISO 639-3 code for Jordanian Sign Language, <jos>, is used as the closest sign language to Syrian Sign Language. Both languages are considered dialects of the Levantine Sign Language with slight vocabulary differences, The final prompt for an “وزارة الصحة” will look like: “<eng> <jos> Ministry of Health”. The prompt is then tokenized with Google’s BERT base uncased, fixed to a sequence length of 256 tokens. The following figure (Figure 7) shows the pipeline of the AraSignCLIP

Fig. 7. A Comprehensive pipeline of AraSignCLIP

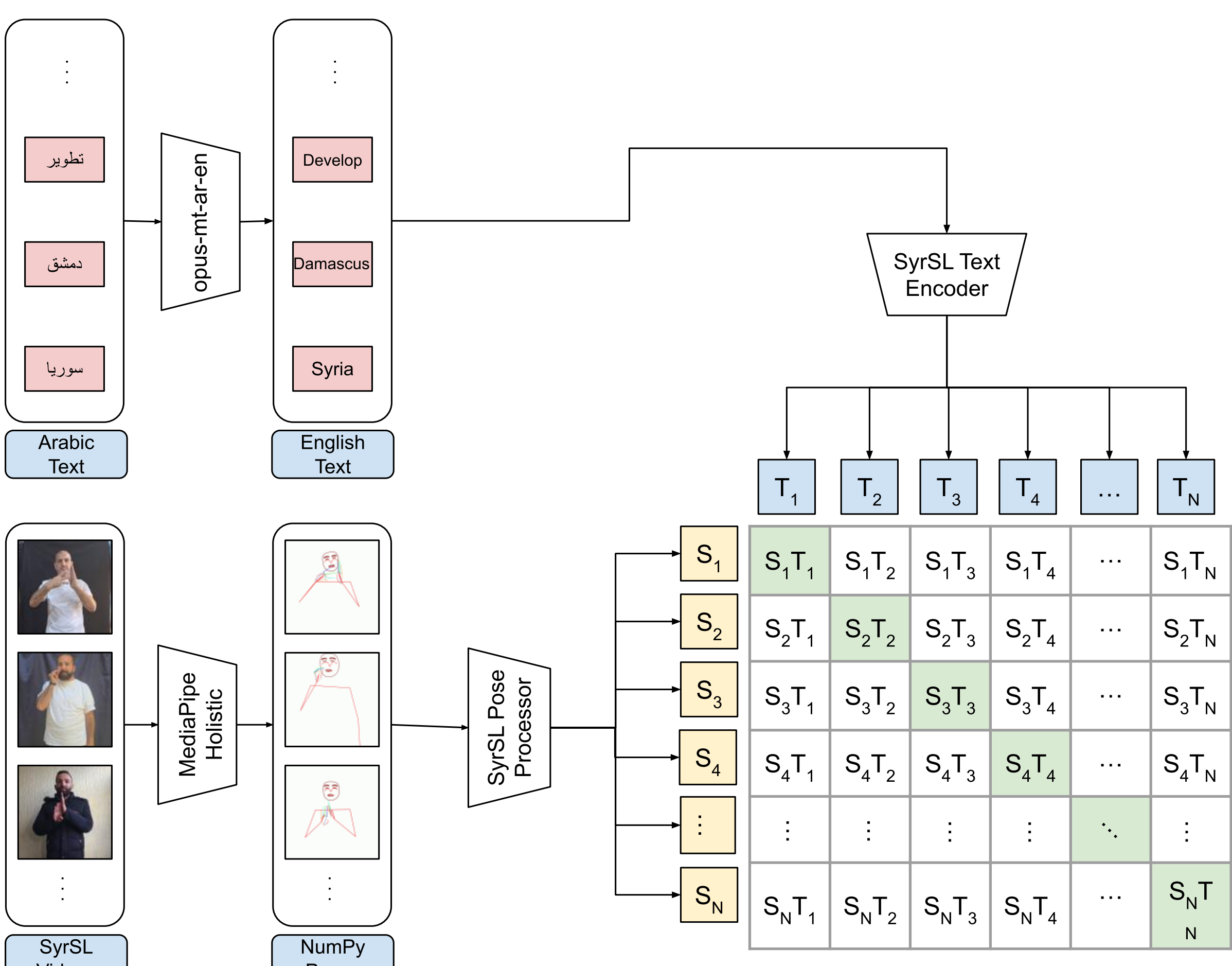

## 4. RESULTS

The dataset was split into training (80%), validation (10%), and testing (10%). It includes recordings from six of the seven signers, where each signer contributes approximately 250 samples corresponding to 50 unique words, each repeated five times. While this setup provides a degree of signer diversity, a fully signer-independent evaluation protocol was not enforced due to the limited number of participants.

*Table 3 Results of finetuning MotionCLIP, T2M-GPT, SignCLIP*

| Model | Component | Best Metric | Result |
|---|---|---|---|
| MotionCLIP | Encoder | Train Loss | 1.3210 |
| | | Validation Loss | 1.0783 |
| | Decoder | Train Loss | 0.7587 |
| | | Validation Loss | 0.5285 |
| T2M-GPT | VQ-VAE (No face features) 168 Window Size | Reconstruction Test Loss | 0.1205 |
| | | Perplexity | 186.69 |
| | | Commit | 0.04445 |
| | GPT (Pose-Delta Weighted) Best VQ-VAE Model | Train Loss | 1.2333 |
| | | Validation Loss | 0.0242 |
| | | Cross-Entropy Test Loss | 0.0242 |
| SignCLIP | - | Training Loss | 1.5264 |
| | | R@5 | 0.5526 |

## 5. DISCUSSION

As shown in Table 3, results indicate that MotionCLIP performed better in retrieval mode because of embedding alignment while generation was unstable, SignCLIP produced decent results for retrieval-based approach due to its reliance on similarity matching rather than full motion synthesis.

SMPL-X (Skinned Multi-Person Linear eXtended) which is a 3D model that represents the human body motion, produced poor results when the hands overlapped, leading to inaccurate hand modeling. In contrast, MediaPipe demonstrated consistent and accurate representations for hands, making it more suitable for our application.

T2M-GPT struggled due to the limited dataset size since Text-to-motion generation is highly data-dependent, the diversity and richness of motion patterns directly affect model generalization. While generative methods such as T2M-GPT show promising qualitative behavior their performance as mentioned is highly dependent on dataset scale, and in this low resource setting retrieval-based approaches demonstrated more stable outputs. However, generative models remain a promising direction for future large-scale SyArSL production along with moving on from word-based to sentence-based datasets while increasing its scale will be more efficient for our research goal. The relatively low cross-entropy loss observed in T2M-GPT is attributed to the limited vocabulary size of motion tokens and constrained dataset diversity, which simplifies the prediction space compared to large-scale benchmarks.

This work operates under significant constraints including limited dataset size and a word-level scope and thus the models are not expected to generalize to a sentence level SLP, or capture full linguistic features especially for non-manual ones such as facial expressions. Additionally, the evaluation is limited to proxy metrics due to the nature of sign language ambiguity. As this study does not include standardized production metrics or human evaluation, comparisons between models should be interpreted with caution.

## 6. Conclusion

In conclusion, our work presented an automated Arabic text-to-Syrian Arabic Sign Language (SyArSL) news-focused translation system, aimed at empowering the deaf and hard of hearing community in Syria. The work evaluated the created dataset using multiple state-of-the-art sign language production approaches, including MotionCLIP, SignCLIP, T2M-GPT. Our dataset, SyriSign, consists of 1500 signed samples for 150 words. It will be publicly available and we hope that it will be an initial resource for further research in SyArSL production and translation. Future work will focus on scaling the dataset to include more signers and shifting to a sentence-level recording, incorporating full facial and hand features to capture full linguistic features, and finally applying standardized evaluation metrics including back translation and human evaluation.

## 7. Acknowledgment

The Authors thank Dr. Mohammad Mohammad for his continuous support and guidance, and for SIGN App Team, Isharati Foundation and Deaf Care Association Foundation in Damascus for their help with recording the dataset

## References

Akkazeh. (2025, January). *حقوق بلا ترجمة: حين يعجز القانون عن سماع الصم*. Retrieved January 18, 2026, from Akkazeh عكازة: https://akkazeh.net/3173/حقوق-بلا-ترجمة-حين-يعجز-القانون-عن-سماع

Alethary, A. A., Aliwy, A. A., & Ali, N. S. (2022). Automated Arabic-Arabic sign language translation system based on 3D avatar technology. *International Journal of Advances in Applied Sciences (IJAAS), 11*, 383–396.

Aulamo, M., Tiedemann, J., Bakshandaeva, D., Boggia, M., Grönroos, S.-A., Nieminen, T., . . . Virpioja, S. (2023). Democratizing Neural Machine Translation with OPUS-MT. *Language Resources and Evaluation*.

Camgöz, N. C., Hadfield, S., & Koller, O. (2018). Neural Sign Language Translation. *IEEE Conf. on Computer Vision and Pattern Recognition.* Salt Lake City, UT.

Grishchenko , I., & Bazarevsky, V. (2020, December 10). *MediaPipe Holistic — Simultaneous Face, Hand and Pose Prediction, on Device.* Retrieved from https://research.google/blog/mediapipe-holistic-simultaneous-face-hand-and-pose-prediction-on-device/

Humanitarian Needs Assessment Programme (HNAP). (2021). Disability in Syria: Investigation on the Intersectional Impacts of Gender, Age and a Decade of Conflict on Persons with Disabilities. Retrieved from https://www.hi-deutschland-projekte.de/lnob/wp-content/uploads/sites/2/2021/09/hnap-disability-in-syria-investigation-on-intersectional-impacts-2021.pdf

Jedlička, P., Krňoul, Z., Železný, M., & Müller, L. (2022). MC-TRISLAN: A Large 3D Motion Capture Sign Language Dataset (Czech Sign Language). *Proceedings of the Twelfth Language Resources and Evaluation Conference (LREC 2022)* (pp. 88–95). Marseille, France: European Language Resources Association (ELRA). Retrieved from https://aclanthology.org/2022.signlang-1.14/

Jiang, Z., Sennrich, R., Moryossef, A., Sant, G., Ebling, S., & M, M. (2024). *SignCLIP: Connecting Text and Sign Language by Contrastive Learning.* Zurich. Retrieved 7 Jan, 2026, from https://arxiv.org/pdf/2407.01264

Koller, O., Zargaran, S., Ney, H., & Bowden, R. (2018). Deep Sign: Enabling Robust Statistical Continuous Sign Language Recognition via Hybrid CNN-HMMs. *International Journal of Computer Vision, 126*(12), 1311–1325. doi:10.1007/s11263-018-1121-3

Latif, G., Mohammad, N., Alghazo, J., AlKhalaf, R., & AlKhalaf, R. (2019). ArASL: Arabic Alphabets Sign Language Dataset. *Data in Brief, 23*. doi:10.1016/j.dib.2019.103777

Li, D., Rodriguez Opazo, C., Yu, X., & Li, H. (2020). Word-level Deep Sign Language Recognition from Video: A New Large-scale Dataset and Methods Comparison. *IEEE/CVF Winter Conference on Applications of Computer Vision (WACV)*, (pp. 1459–1468). doi:10.1109/WACV45572.2020.9093305

Lugaresi, C., Tang, J., Nash, H., McClanahan, C., Uboweja, E., Hays, M., . . . Grundmann, M. (2019). *MediaPipe: A Framework for Building Perception Pipelines.* arXiv.

Luqman, H., Mohandes, M., Mahmoud, S., & I. Sidig, A. A. (2021). KArSL: Arabic Sign Language Database. *ACM Transactions on Asian and Low-Resource Language Information Processing*, 1 - 19. Retrieved 1 19, 2026, from https://doi.org/10.1145/3423420

Nagoudi, E. M., Elmadany, A., & Abdul-Mageed, M. (2022). AraT5: Text-to-Text Transformers for Arabic Language Generation. *ACL 2022*. Retrieved from https://arxiv.org/abs/2109.12068

Saunders, B., Camgoz, N. C., & Bowden, R. (2020). Progressive Transformers for End-to-End Sign Language Production. *ECCV 2020*. Retrieved from https://arxiv.org/abs/2004.14874

Tevet, G., Gordon, B., Hertz, A., Bermano, A. H., & Cohen-Or, D. (2022). MotionCLIP: Exposing Human Motion Generation to CLIP Space.

Vaezi Joze, H. R., & Koller, O. (2019). MS-ASL: A Large-Scale Data Set and Benchmark for Understanding American Sign Language. *British Machine Vision Conference (BMVC).* doi:10.48550/arXiv.1812.01053

Zelinka, J., & Kanis, J. (2020). Neural Sign Language Synthesis: Words Are Our Glosses. *Conference: 2020 IEEE Winter Conference on Applications of Computer Vision (WACV)*, (pp. 3395–3403). doi:10.1109/WACV45572.2020.9093516

Zhang, J., Zhang, Y., Cun, X., Huang, S., Zhang, Y., Zhao, H., . . . Shen, X. (2023). *T2M-GPT: Generating Human Motion from Textual Descriptions with Discrete Representations.* arXiv. CVPR 2023.

Zifan Jiang, G. S. (2024). SignCLIP: Connecting Text and Sign Language by Contrastive Learning.